\documentclass{article}

\usepackage{arxiv}
\usepackage[utf8]{inputenc} % Allow utf-8 input
\usepackage[T1]{fontenc} % Use 8-bit T1 fonts
\usepackage{hyperref} % Hyperlinks
\usepackage{url} % Simple URL typesetting
\usepackage{booktabs} % Professional-quality tables
\usepackage{amsfonts} % Blackboard math symbols
\usepackage{nicefrac} % Compact symbols for 1/2, etc.
\usepackage{microtype} % Microtypography
\usepackage{graphicx}
\usepackage{doi}

\title{Multi-head automated segmentation by incorporating detection head into the contextual layer neural network }

\author{
  Edwin Kys \\
  Researcher \\
  Linnear  \\
  Austin, Texas \\
  \texttt{edwin@linnearai.com}
  \And
  Febian Febian\\
  Researcher   \\
  UCL / Linnear \\
  London, United Kingdom \\
  \texttt{febian.febian.19@ucl.ac.uk}
}

\renewcommand{\headeright}{}
\renewcommand{\shorttitle}{Multi-head Contextual Layer Neural Network}

\hypersetup{
  pdftitle={Multi-head Contextual Layer Neural Network},
  pdfsubject={Medical imaging and Artificial Intelligence},
  pdfauthor={Edwin and Febian},
  pdfkeywords={CT, AI, CNN, DL, Swin U-Net},
}

\begin{document}

\maketitle{}

% Abstract
\begin{abstract}
  Deep-learning–based auto-segmentation is increasingly used in radiotherapy, but conventional models often produce anatomically implausible false positives, or “hallucinations,” in slices lacking target structures. We propose a gated multi-head Transformer architecture based on Swin U-Net, augmented with inter-slice context integration and a parallel detection head, which jointly performs slice-level structure detection via a multi-layer perceptron and pixel-level segmentation through a context-enhanced stream. Detection outputs gate the segmentation predictions to suppress false positives in anatomically invalid slices, and training uses slice-wise Tversky loss to address class imbalance. Experiments on the Prostate-Anatomical-Edge-Cases dataset from The Cancer Imaging Archive demonstrate that the gated model substantially outperforms a non-gated segmentation-only baseline, achieving a mean Dice loss of 0.013 ± 0.036 versus 0.732 ± 0.314, with detection probabilities strongly correlated with anatomical presence, effectively eliminating spurious segmentations. In contrast, the non-gated model exhibited higher variability and persistent false positives across all slices. These results indicate that detection-based gating enhances robustness and anatomical plausibility in automated segmentation applications, reducing hallucinated predictions without compromising segmentation quality in valid slices, and offers a promising approach for improving the reliability of clinical radiotherapy auto-contouring workflows.
\end{abstract}

% Keywords
% All keywords should be written in Title Case.

\section{Introduction}
\label{sec:intro}
Computed tomography (CT) imaging plays a central role in modern clinical workflows, particularly in radiotherapy treatment planning, where accurate delineation of target volumes and organs-at-risk (OARs) is essential for optimal dose delivery and minimization of normal tissue complications. Manual contouring or segmentation of anatomical structures on CT is widely regarded as the clinical standard; however, it remains a time-consuming, labor-intensive process that is subject to both inter- and intra-observer variability \cite{bibault2023DLnarrativereview}. Deep learning–based auto-contouring seeks to address these limitations by providing rapid, reproducible segmentation of structures directly from CT volumes.

Auto-segmentation methods in medical imaging have evolved substantially over the past decade. Traditional approaches, such as atlas-based registration and deformable models, laid early groundwork for algorithmic contouring but were often challenged by the considerable anatomical variability, imaging artifacts, and limited soft-tissue contrast inherent in CT imaging \cite{Wachinger2015ContourDrivenAtlas}. The advent of convolutional neural networks (CNNs) and related deep learning architectures ushered in significant performance gains, with models now routinely achieving expert-level segmentation accuracy in many clinical contexts. A particularly influential architecture in this domain is the U-Net, which couples an encoder–decoder structure with skip connections to capture both global context and fine spatial details for segmentation tasks \cite{Ronneberger2015UNet}. A study by Ng et al.(2022) illustrates significant performance gains of deep learning-based model compared to the atlas-based one. In a retrospective comparison involving 45 head and neck radiotherapy patients, the deep learning–based segmentation method consistently outperformed the atlas-based approach by achieving median Dice similarity coefficients above 0.80 for key organs such as the brainstem, mandible, eyes, and spinal cord, while also exhibiting less variability and requiring less time for contouring than the atlas-based method \cite{Ng2022Clinical_DLAutoContouring}.

Deep learning–based auto-segmentation has been applied across a range of CT-based radiotherapy scenarios, demonstrating substantial reductions in contouring time and variability when compared with manual delineation. For example, reviews of such methods have shown that deep learning auto-contours can achieve accuracy comparable to manual expert contours across multiple sites and structures while drastically reducing user time \cite{Moran2025EvaluatingDeepLearning}. Another study by Zhu et al. (2018) suggests a 3D squeeze-and-excitation U-Net architecture trained on 261 head and neck CT volumes, and demonstrated that it improved average Dice similarity coefficients by approximately 3.3\% compared with the previous MICCAI 2015 head-and-neck segmentation benchmark while processing an entire organ-at-risk set in roughly 0.12 seconds in a single forward pass \cite{Zhu2018DLvolumetric}. These methods also show promise in reducing inter-observer variability, one of the key bottlenecks in clinical contouring workflows, and in maintaining consistency when applied to independent datasets \cite{Wong2020ComparingDeepLearning, Arjmandi2025EvaluatingDosimetricImpact}.

Despite these advances, several challenges remain. CT images are inherently limited in soft-tissue contrast, which can confound boundary definition for certain organs or targets \cite{Weiss2010CBCTSoftTissue}. Additionally, generalization across imaging protocols, institutions, and patient populations remains a barrier to broad clinical adoption \cite{Suleman2025AIGeneralizabilityRadiology}. Many deep learning approaches also require large volumes of high-quality annotated data, which are costly to curate and may not be uniformly available across all anatomical sites or treatment settings \cite{Wang2021AnnotationEfficient}. Robust uncertainty estimation and clinically acceptable performance thresholds are further areas of active investigation.

In this work, we propose a model that incorporates contextual layer neural network integrated with detection head for auto-contouring on CT images. The proposed approach is designed to reduce hallucination, hence, improving the volumetric dice performance thereby improving contour quality and reducing the amount of manual correction required by clinicians. The work is trained and evaluated on a public dataset provided by TCIA namely Prostate-Anatomical-Edge-Cases collection and compare against established baselines using standard quantitative metrics and qualitative expert review.

\section{Materials and Methods}
\label{sec:method}
Medical image segmentation has seen substantial advances with the adoption of deep learning, particularly CNN and, more recently, Transformer-based architectures. Models such as U-Net and its numerous variants have become the de facto standard due to their strong performance in pixel-wise prediction tasks. However, a persistent challenge in volumetric and slice-based medical imaging is the phenomenon commonly referred to as hallucination, where models predict anatomically implausible structures in image slices where those structures do not exist. This issue is especially pronounced in organs with limited spatial extent along a given axis, such as lungs near the apex or base, small tumours, or vessels that appear intermittently across slices \cite{zhou2024deeplearningsegmentationsmallvolumes}. 
Hallucination arises primarily because traditional segmentation models are optimized solely for pixel-level accuracy, without an explicit mechanism to reason about structure presence or absence at the slice-level. As a result, models may extrapolate learned spatial patterns beyond their valid anatomical context, leading to false positive segmentations that are clinically misleading \cite{rickmann2023haloshallucinationfreeorgansegmentation}.

The emerging transformer-based models, including Swin U-Net variants, have been introduced to address limitations of CNNs in capturing long-range dependencies. Hierarchical Transformers leverage self-attention mechanisms to model global context while maintaining computational efficiency through windowed attention \cite{Zhang2025MedicalImageSegmentation}. In medical imaging, these architectures have demonstrated improved boundary delineation and robustness to noise.
Despite these improvements, Transformers alone do not inherently solve the hallucination problem. While temporal or inter-slice context can reduce inconsistency as presented in a study by An (2025) \cite{An2025Sli2VolPlus}, excessive reliance on contextual cues may actually exacerbate hallucinations by encouraging the model to propagate structures across slices where they are anatomically absent. This limitation motivates architectures that explicitly separate structure existence reasoning from pixel-wise segmentation.

Multi-task learning has been widely explored in medical imaging as a means to improve generalization by sharing representations across related tasks. Recent work has explored architectures that combine detection and segmentation sequentially. Detection–segmentation pipelines often use detection outputs to define regions of interest for segmentation, thereby reducing false positives \cite{Araujo_2018Objectdetextionxsegmentation}. However, such pipelines are typically staged and do not benefit from fully shared representations.

\begin{figure}
    \centering
    \includegraphics[width=\linewidth]{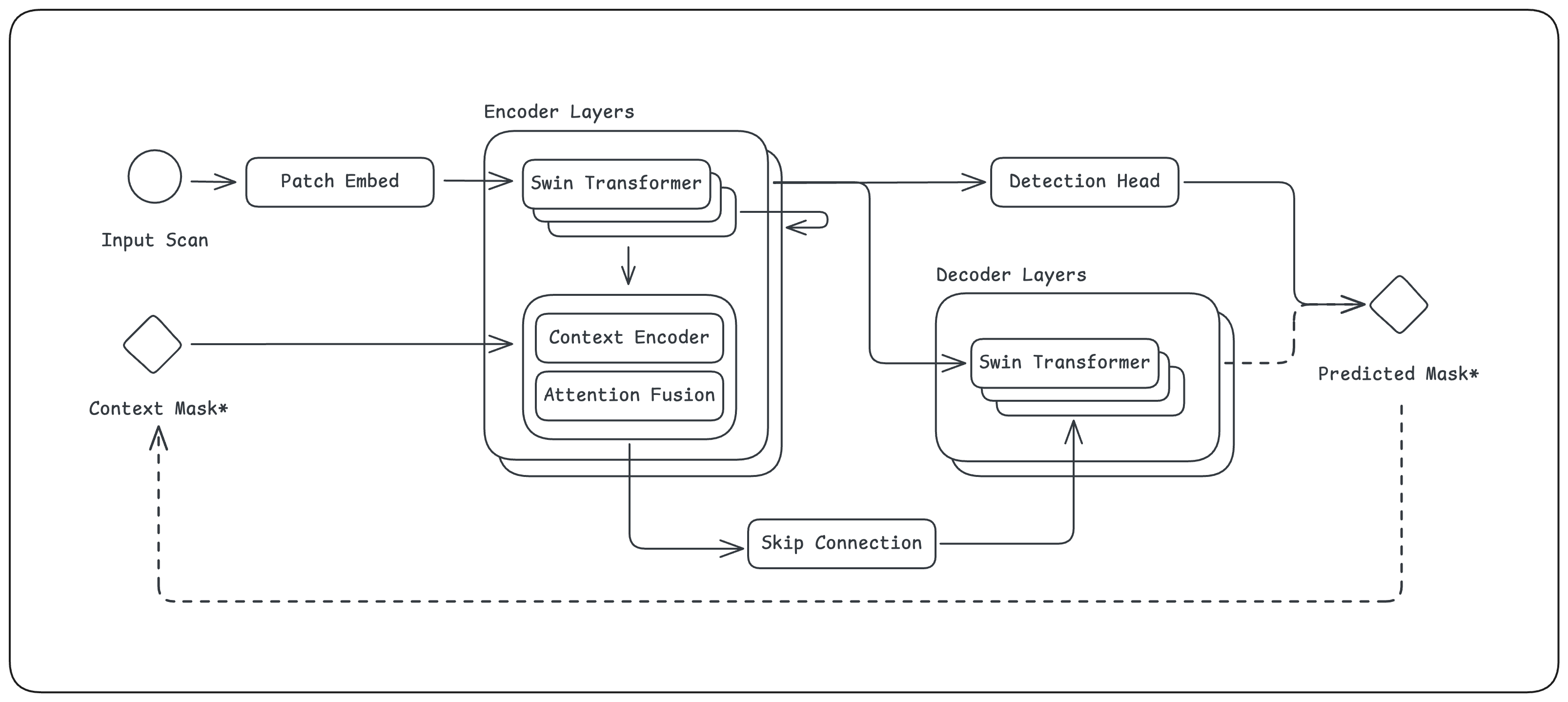}
    \caption{Overview of N2 model workflow consisting of patch embedding, encoder layers with context integration, skip connections to the decoder, and temporal context fusion with the addition of detection head in its pipeline.}
    \label{fig:n2v1workflow}
\end{figure}

Our proposed model, N2 architecture as shown in Figure \ref{fig:n2v1workflow}, builds on this paradigm by introducing dual processing streams in parallel: a context-free detection stream and a context-enhanced segmentation stream. The segmentation stream is based on Swin U-Net and enhanced with temporal context integration. Through a cross-attention mechanism, N2 incorporates previous segmentation masks into the encoder, which strengthens its ability to handle inter-frame variations and promotes consistent segmentations over time. This design choice highlights segmentation benefits from temporal or contextual fusion whereas detection should rely primarily on the current image content to avoid biased presence decisions driven by neighbouring slices.
In this model, detection outputs can be used to gate or modulate segmentation predictions. Soft gating strategies scale segmentation probabilities by detection confidence, while hard gating applies binary masks based on detection thresholds. These mechanisms are especially effective for suppressing hallucinated segmentations without degrading performance on slices where the structure truly exists which will be presented in more detail in later section.

The N2 workflow starts with an initial patch-embedding layer that converts the raw input tensor of shape (B, CS, H, W), where B is batch size, CS is the number of channels, and H, W are spatial dimensions, into a structured sequence of tokens by partitioning the image into non-overlapping patches of size (PH, PW), flattening them, and projecting them into a D-dimensional embedding space using a linear or convolutional projection. This produces a feature tensor of shape (B, N, D) that enables efficient capture of local spatial patterns while reducing the computational cost compared to processing the whole image. The encoder then builds hierarchical feature representations by passing these patch tokens through multiple stages of Swin Transformer blocks that apply window-based and shifted window self-attention, interleaved with patch-merging layers that downsample spatial resolution and increase channel depth, all while supporting multi-scale feature extraction. Skip connections link encoder and decoder stages to preserve fine spatial details. 

Simultaneously, the output features from the encoder are routed into a multi-head processing module: one stream feeds the segmentation decoder, while a parallel detection stream leads into a detection head implemented as a multi-layer perceptron (MLP) classifier to predict structure presence or absence from the encoded representations. In the decoder, patch-expanding layers upsample the features progressively, merging them with corresponding encoder outputs via skip connections to restore spatial resolution and recover both contextual and fine structural information effectively. This dual-head design enables the model to both delineate anatomical structures and determine their existence within a given input, improving overall reliability. Finally, normalized features are linearly projected to per-class segmentation logits, reshaped into spatial grids and upsampled back to the original resolution to produce dense, pixel-level segmentation outputs.

\begin{table*}[h]
\centering
\caption{End-to-end example flow of the N2 architecture input and output at each stage.}
\label{tab:N2_pipeline}
\begin{tabular}{@{}cllll@{}}
\toprule
Stage & Description & Input Shape & Output Shape & Notes \\ \midrule
1 & Patch Embedding & $(8, 1, 256, 256)$ & $(8, 4096, 96)$ & Patch Size $(4,4)$, $D=96$ \\
\midrule
2.1 & Encoder Stage 1 & $(8, 4096, 96)$ & $(8, 1024, 192)$ & Channels=192, Tokens=1024 \\
& Temporal Context Fusion & $(8, 1, 256, 256)$ & $(8, 1024, 192)$ & Context Encoder \\
 &  & + $(8, 1024, 192)$ &  & + Attention Fusion \\
2.2 & Encoder Stage 2 & $(8, 1024, 192)$ & $(8, 256, 384)$ & Channels=384, Tokens=256 \\
& Temporal Context Fusion & $(8, 1, 256, 256)$ & $(8, 256, 384)$ & Context Encoder \\
 &  & + $(8, 256, 384)$ &  & + Attention Fusion \\
2.3 & Encoder Stage 3 & $(8, 256, 384)$ & $(8, 64, 768)$ & Channels=768, Tokens=64 \\
& Temporal Context Fusion & $(8, 1, 256, 256)$ & $(8, 64, 768)$ & Context Encoder \\
 &  & + $(8, 64, 768)$ &  & + Attention Fusion \\
2.4 & Encoder Stage 4 & $(8, 64, 768)$ & $(8, 64, 768)$ & Channels=768, Tokens=64 \\
 & Temporal Context Fusion & $(8, 1, 256, 256)$ & $(8, 64, 768)$ & Context Encoder \\
 &  & + $(8, 64, 768)$ &  & + Attention Fusion \\
\midrule
3.1 & Detection Head & $(8, 64, 768)$ & $(8, 3)$ & ROI Channel=3 \\
3.1 & Decoder Stage 4 & $(8, 64, 768)$ & $(8, 256, 384)$ & Channels=384, Tokens=256 \\
3.2 & Decoder Stage 3 & $(8, 256, 384)$ & $(8, 1024, 192)$ & Channels=192, Tokens=1024 \\
3.3 & Decoder Stage 2 & $(8, 1024, 192)$ & $(8, 4096, 96)$ & Channels=96, Tokens=4096 \\
3.4 & Decoder Stage 1 & $(8, 4096, 96)$ & $(8, 4096, 96)$ & Channels=96, Tokens=4096 \\
\midrule
4.1 & Normalization & $(8, 4096, 96)$ & $(8, 4096, 96)$ & LayerNorm \\
4.2 & Linear Projection & $(8, 4096, 96)$ & $(8, 4096, 1)$ & Segmentation Classes=1 \\
4.3 & Reshape & $(8, 4096, 1)$ & $(8, 1, 64, 64)$ & Tokens $\to$ Grid \\
4.4 & Upsample & $(8, 1, 64, 64)$ & $(8, 1, 256, 256)$ &  \\ 
\bottomrule
\end{tabular}
\end{table*}

In this work, we utilized the publicly accessible Prostate-Anatomical-Edge-Cases dataset from The Cancer Imaging Archive (TCIA), which includes CT scans with accompanying RTSTRUCT segmentations from 131 individuals diagnosed with prostate cancer, totaling roughly 17 GB of data \cite{Thompson2023PelvicAutosegmentationEdgeCases}. The collection originates from a single-institution, retrospective review of 950 consecutive prostate adenocarcinoma patients treated with definitive radiotherapy between 2011 and 2019, from which 112 cases exhibiting notable anatomical variations—such as hip arthroplasty, enlarged median prostate lobes, “droopy” seminal vesicles, and the presence of urinary catheters—were identified as edge cases, alongside 19 randomly selected normal cases for comparison. For every subject, expert manual segmentations were provided for the prostate, rectum, bladder, and bilateral femoral heads, establishing a robust ground truth for treatment planning and algorithm evaluation.

Medical image segmentation often suffers from severe class imbalance, particularly when structures occupy a small fraction of the image. To train the proposed model, we used Tversky loss, a generalization of overlap-based metrics that introduces tunable weights to explicitly balance penalties for false positives and false negatives, thereby enhancing boundary precision making it well-suited for segmentation tasks where missing a structure is more costly than slight over-segmentation \cite{Terven2025LossFunctionsMetrics}. The loss was computed on a slice-by-slice basis so that each 2D slice contributed equally to training, preventing larger structures from disproportionately influencing the objective.

To improve model generalization and robustness, we applied a suite of preprocessing and data augmentation steps during training. Image intensities were first standardized using Min–Max scaling, and spatial variability was addressed through random 90° rotations and horizontal and vertical flips. Elastic deformations were incorporated to mimic realistic anatomical variability, while photometric variations such as brightness and contrast perturbations, along with controlled Gaussian noise injection, were used to simulate diverse acquisition conditions. Finally, all image volumes were normalized to zero mean and unit variance. This combination of geometric, photometric, and noise augmentations created a varied training distribution while preserving the anatomical fidelity necessary for accurate medical image segmentation.

The model was trained for 100 epochs with a batch size of 8 using the AdamW optimizer, a variant of Adam that separates weight decay from the gradient update to improve regularization, with an initial learning rate of $1 \times 10^{-5}$. The network has about 110 million trainable parameters and was implemented in PyTorch. Training was performed on a single NVIDIA RTX 4000 Ada GPU with 20 GB of VRAM, supported by a host machine equipped with 50 GB of RAM and 9 VCPUs.

\section{Results and Discussion}
\label{sec:results}
Quantitative evaluation was conducted using slice-wise Dice loss and detection probability outputs to compare the proposed gated multi-head architecture with the corresponding non-gated segmentation-only baseline. A total of 150 axial slices were analyzed for each model.

\begin{figure}
    \centering
    \includegraphics[width=0.7\linewidth]{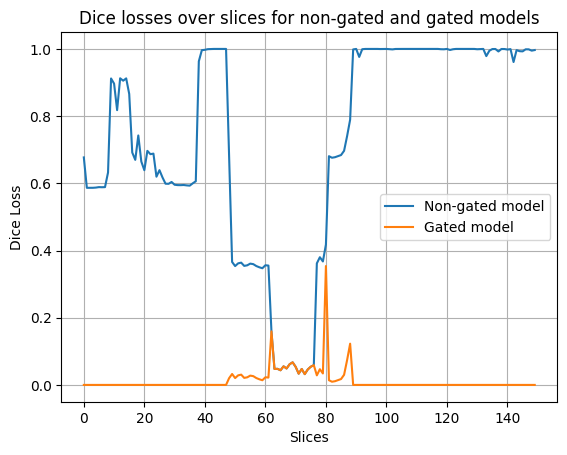}
    \caption{Dice loss comparison between the gated multi-head model and the non-gated segmentation-only model. }
    \label{fig:dicelossresult}
\end{figure}

The gated multi-head model demonstrated a substantial improvement in segmentation performance, exhibiting a markedly lower mean volumetric Dice loss of 0.013 ± 0.036 compared with 0.732 ± 0.314 for the non-gated configuration. In addition to the lower average loss, the gated model showed significantly reduced variability across slices, indicating more stable and consistent performance. By contrast, the non-gated model displayed pronounced fluctuations in Dice loss across all slices, particularly in slices near anatomical boundaries, reflecting an increased incidence of false positive segmentations. These trends are quantitatively summarized in Figure \ref{fig:dicelossresult}.

\begin{figure}
    \centering
    \includegraphics[width=0.85\linewidth]{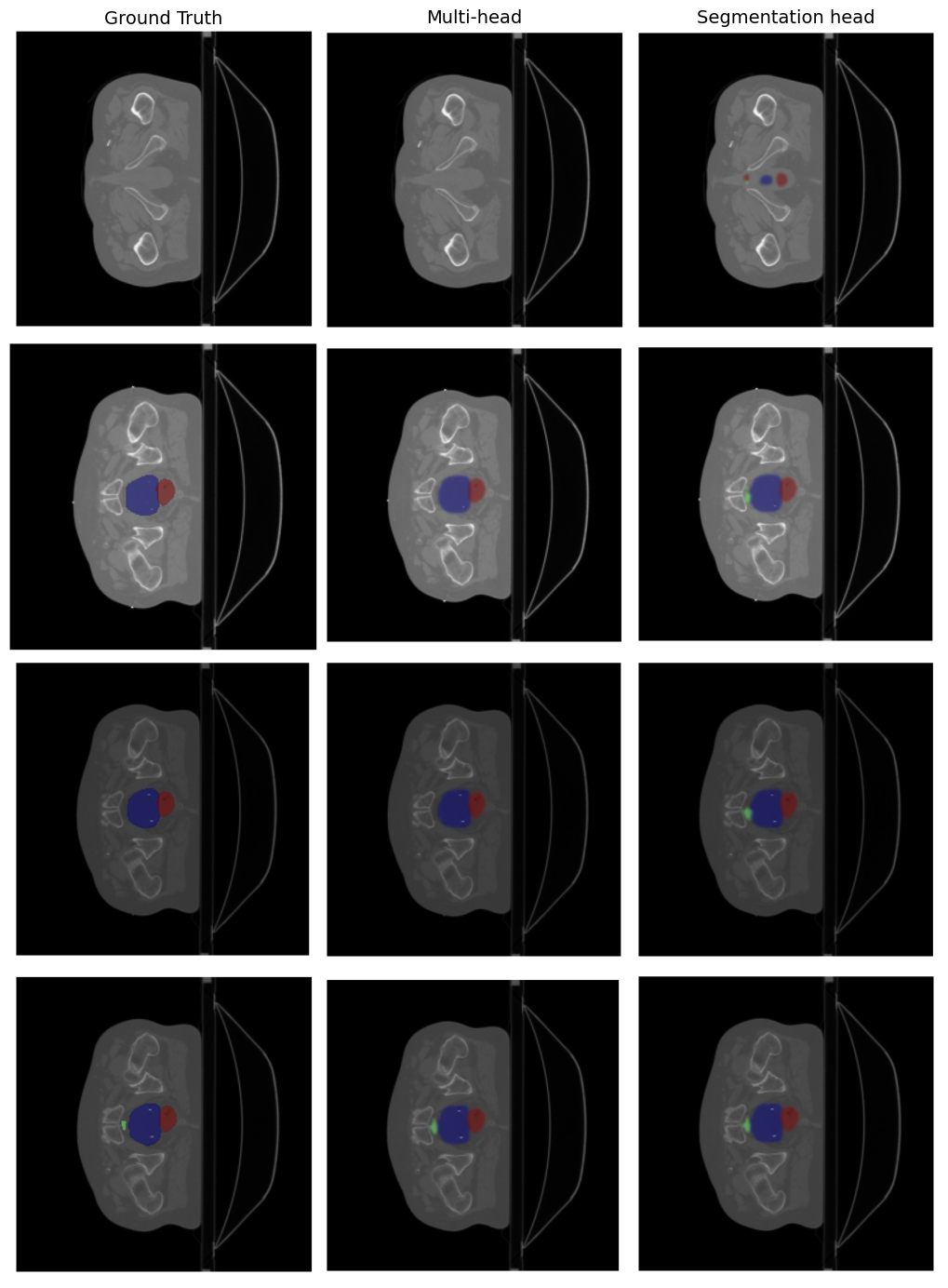}
    \includegraphics[width=0.5\linewidth]{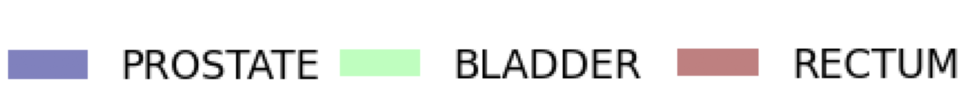}
    \caption{Comparison of multi-channel segmentation results for prostate, bladder, and rectum across sequential slices (45, 60, 61, and 62) from a representative case. Ground truth annotations are shown in the left column, predictions from the gated multi-head model in the middle column, and predictions from the segmentation-only model in the right column.}
    \label{fig:continuousslicesresult}
\end{figure}

Analysis of the detection probability outputs further supports these findings. In the gated architecture, the detection head consistently assigned near-zero confidence scores to slices where the target anatomy was absent. This behaviour effectively suppressed segmentation predictions via the gating mechanism, resulting in near-zero Dice loss in non-anatomical slices. In contrast, the non-gated model produced segmentation outputs regardless of anatomical presence, leading to persistently elevated Dice loss values in slices where no structure should be present. This behaviour highlights the inherent limitation of segmentation-only architectures in handling anatomically absent regions.

Qualitative results across continuous slices are illustrated in Figure \ref{fig:continuousslicesresult}. In the first row, the non-gated segmentation-only model exhibits clear hallucinated segmentations, whereas the gated multi-head model correctly suppresses segmentation, in agreement with the ground truth. Across subsequent rows, hallucinations persist in the non-gated model, including spurious bladder segmentations that propagate across sequential slices. These false positives emphasize the tendency of segmentation-only models to produce anatomically implausible predictions when contextual awareness is lacking. When the bladder structure is finally present in the ground truth, the non-gated model achieves segmentation quality comparable to the gated approach. Refer back to the Figure \ref{fig:dicelossresult}, the dice loss of non-gated model overlaps with the gated model showcasing the effectiveness of the baseline model architecture relying on contextual layer integrated Swin U-Net model. However, this occurs only after a prolonged sequence of erroneous predictions. 

Representative multi-channel segmentation results for prostate, bladder, and rectum are shown in Figure \ref{fig:continuousslicesresult}. The gated multi-head model consistently aligns with the ground truth across both anatomical and non-anatomical slices, while the segmentation-only model demonstrates false positives in regions where structures are absent. Together, these qualitative and quantitative findings indicate that detection-based gating provides a clear advantage, yielding lower Dice loss, reduced variance, and improved anatomical plausibility. These improvements are particularly relevant for clinical radiotherapy applications involving structures with limited spatial extent, where conventional non-gated segmentation models are especially prone to hallucinated predictions.

% \section{Discussion}
% \label{sec:discussion}
% \input{section_src/discussion}

\section{Conclusion}
\label{sec:conclusion}
This work demonstrates that hallucination in medical image segmentation is not merely a consequence of data limitations or optimization challenges, but a fundamental structural limitation of models trained exclusively for pixel-level accuracy without explicit modeling of anatomical presence. By introducing a dual-head architecture that decouples slice-level structure detection from pixel-wise segmentation and incorporates detection-based gating, the proposed approach effectively suppresses false positive predictions in anatomically invalid slices.

The non-gated segmentation-only model produced persistent false positive predictions in anatomically absent regions, resulting in a high mean volumetric Dice loss of 0.732 ± 0.314. In contrast, incorporating a multi-head architecture that combines a slice-level detection component based on an MLP with a contextual Swin U-Net segmentation backbone effectively mitigates this hallucination behaviour. The proposed multi-head model achieves a substantially lower mean volumetric Dice loss of 0.013 ± 0.036, along with markedly reduced variance across slices. This numerical improvement is particularly pronounced in boundary and edge-case slices, where hallucinated predictions dominate the error profile of the non-gated baseline. The detection head reliably assigns near-zero confidence to slices lacking target anatomy, preventing the propagation of erroneous segmentations while preserving accurate predictions when structures are present.

Overall, these results provide strong quantitative evidence that detection-aware, multi-head architectures significantly enhance anatomical plausibility and robustness in deep learning–based auto-segmentation. By addressing hallucination at a structural level, rather than relying solely on pixel-wise optimization, the proposed framework offers a promising and clinically meaningful advancement toward safer and more reliable radiotherapy segmentation workflows. Future work will focus on extending this approach to additional anatomical sites, validating generalizability across institutions, and exploring uncertainty-aware gating mechanisms to further improve clinical safety.

% Referencing handler using IEEE style
\bibliographystyle{ieeetr}
\bibliography{references}

\end{document}